\documentclass[journal]{IEEEtran}

\usepackage{graphicx}
\usepackage{pdfpages}
\usepackage{cite} 
\usepackage{amsmath}
\usepackage{amsfonts}
\usepackage{bm}
\usepackage{booktabs}
\usepackage{multirow}
\usepackage{mathtools}
\usepackage{amssymb}
\usepackage{pifont}
\usepackage{url}
\usepackage{xcolor}
\usepackage{hyperref}
\usepackage{threeparttable}
\usepackage{optidef}
\usepackage{algorithmicx}
\usepackage[ruled]{algorithm}
\usepackage[noend]{algpseudocode}
\usepackage{subcaption}
\usepackage{tabularx}

\newcommand{\etal}{\textit{et~al.}}
\usepackage{amsthm}
\theoremstyle{plain}

\theoremstyle{definition}

\theoremstyle{remark}

\begin{document}

\title{Force-Driven Hook Disassembly via Constraint-Aware Object-Centric Reinforcement Learning}

\author{Nadeen~Elhadad, Takuya~Kiyokawa, and~Kensuke~Harada%
\thanks{This work has been submitted to the IEEE for possible publication. Copyright may be transferred without notice, after which this version may no longer be accessible.}%
\thanks{*Corresponding Author: Nadeen Elhadad. (e-mail: \texttt{nadeenelhadad99@gmail.com})}%
\thanks{N. Elhadad, T. Kiyokawa, and K. Harada are with the Department of Systems Innovation, Graduate School of Engineering Science, The University of Osaka, 1-3 Machikaneyama, Toyonaka, Osaka, Japan.}%
\thanks{K. Harada is with the Industrial Cyber-physical Systems Research Center, The National Institute of Advanced Industrial Science and Technology (AIST), 2-3-26 Aomi, Koto-ku, Tokyo, Japan.}%
}

\markboth{Submitted to IEEE Transactions on Automation Science and Engineering}%
{Elhadad \etal: Force-Driven Hook Disassembly via Constraint-Aware Object-Centric Reinforcement Learning}

\maketitle

\begin{abstract}
The growing volume of end-of-life products has intensified the need for automated disassembly, yet hook-type connections remain challenging: their orientations vary, they deform under load, and their geometry is hidden, so vision alone cannot identify a feasible non-destructive disassembly direction. This work proposes a constraint-aware reinforcement learning framework that infers feasible extraction directions online from contact forces. The policy is trained in a robot-agnostic, object-centric simulation on a simplified peg-in-hole pair and deployed zero-shot on a physical robot, where it transfers to real connections with unseen, multi-stage constraint structures. Because instantaneous contact forces are noisy and locally ambiguous, the policy is coupled with a particle filter belief that provides the policy with temporally integrated directional information and reduces redundant exploratory motion. The approach is validated on four real-world tasks, from peg-in-hole pairs to the inner propeller and front cover of an end-of-life air-conditioning unit. The full method achieves success rates of $100\%$, $100\%$, $90\%$, and $75\%$ with the highest path efficiency on every task, and raises the success rate on the air-conditioning cover from $45\%$ to $75\%$ relative to a policy trained without the belief. 
These results indicate that force-driven reinforcement learning can enable disassembly without product-specific models or demonstrations.

\end{abstract}


\def\abstractname{Note to Practitioners}
\begin{abstract}
This work was motivated by the need to automate disassembly of end-of-life products such as outdoor air-conditioning units for material recovery and reuse. Their hook-type connections are hidden inside the housing and vary between units, preventing camera-based systems and pre-programmed routines from determining which direction a panel should be pulled without breaking it.

This work proposes a robot control approach that relies on the forces felt by the robot during interaction, similar to how a human unfastens a hidden clip. A reinforcement learning policy, trained entirely in low-cost simulation on a simplified peg-in-hole task, interprets these resistance signals and adjusts its motion accordingly, while an estimator maintains a running belief of the most likely release direction to keep the motion decisive. Because training relies only on a generic assembly, the approach avoids building a disassembly trajectory or CAD model for every product variant.

The approach was validated on the inner propeller and the front cover of a previously unseen end-of-life air-conditioning unit, as well as on two peg-in-hole pairs, without additional training. Compared with a hand-designed force-following rule using the same force signal, it moved parts more directly on every task and remained reliable where the release direction changes partway through extraction, where the rule often failed; the rule remains a reasonable choice for simple single-stage connections, which it completed reliably but more slowly.

However, the method requires a clear force signal, and the grasp location is currently chosen manually; a poorly chosen grasp can reduce reliability. Future work targets automated grasp selection and improved handling of low-information contact states.
\end{abstract}

\section{Introduction}
Electronic waste (e-waste) refers to any discarded device that operates using a plug or battery. Driven by rapid technological advancement, increased consumption, and shorter product lifespans, global e-waste generation increased by 82\% between 2010 and 2022. In 2022 alone, e-waste reached 62 million tonnes and is projected to rise further to approximately 82 million tonnes by 2030~\cite{balde2024global}.

To address this challenge, increasing efforts have been directed toward remanufacturing to enable the recovery of materials from end-of-life (EoL) products for reuse in new production cycles, thereby supporting sustainable manufacturing practices.
To mitigate labor shortages, automation and human-robot collaboration of disassembly processes has become increasingly important\cite{lee2024review}. In this context, the robotic disassembly of tightly constrained connections, such as internal hook mechanisms with unseen geometries, is essential for effective remanufacturing. 

\begin{figure}[t]
    \centering
    \includegraphics[width=\linewidth]{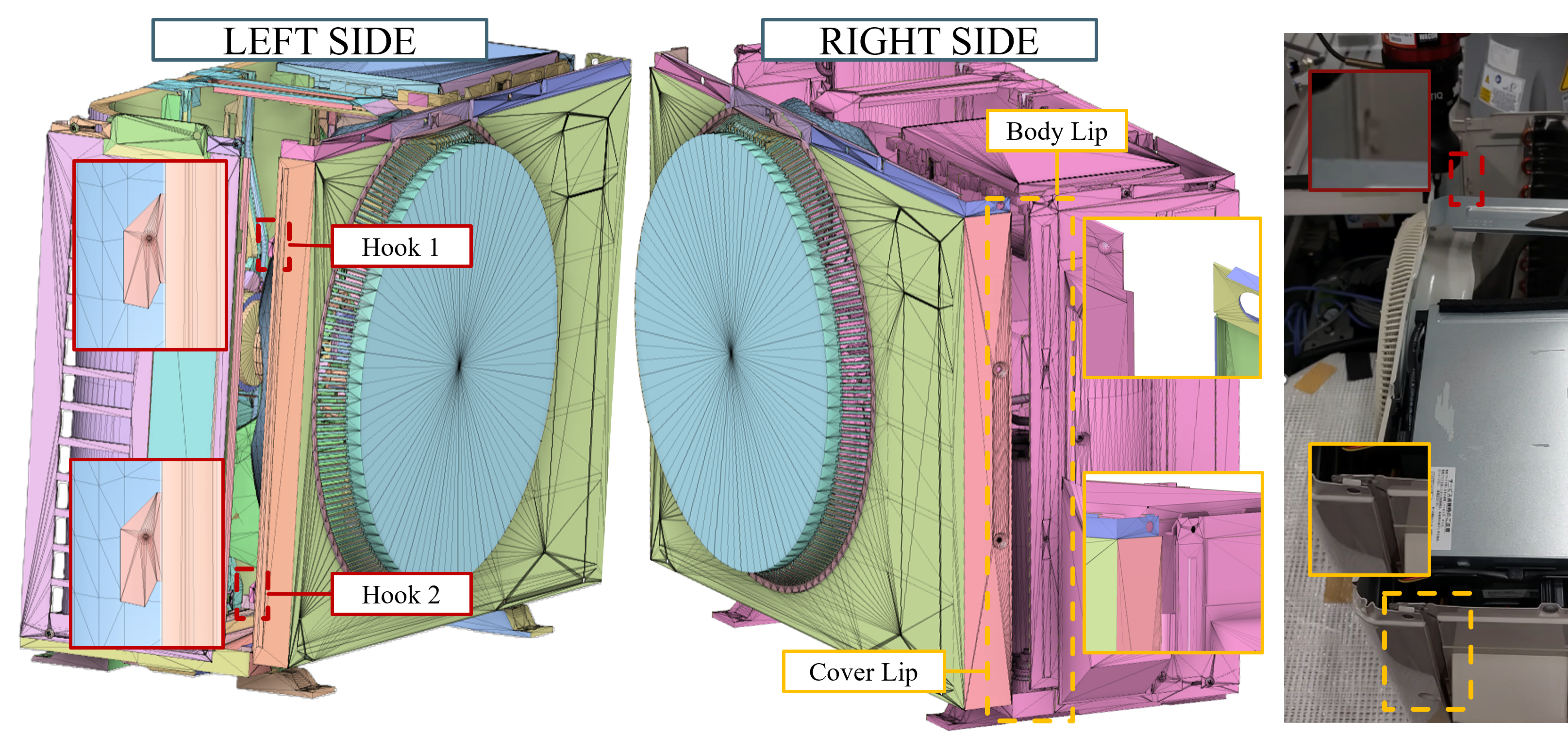}
    \caption{CAD render of the target outdoor air-conditioning unit viewed from two angles. The front cover is retained by two asymmetric constraint mechanisms: two downward-facing hooks on the left edge and two lip-type fasteners on the right edge. The cover material exhibits mild elastic compliance, permitting local elastic deformation that influences the feasible extraction path.}
    \label{fig:aircon_constraintswithreal}
\end{figure}

As a representative case study, we consider the front cover of the commercial outdoor air-conditioning unit in Fig.~\ref{fig:aircon_constraintswithreal}. The cover is retained by two dissimilar constraint types: two downward hooks on one side and two lip-style fasteners on the opposite side. This mixed fastening configuration introduces distinct geometric constraints on the disassembly motion, motivating its treatment as a constraint-aware manipulation problem. Furthermore, the cover exhibits mild structural compliance, allowing local elastic deformation under load. Consequently, the feasible extraction path depends on the interaction between the cover's elastic deformation and the coupled constraint mechanisms rather than a single fixed geometric direction. The combined actions of these two constraint types produce a coupled multi-stage disassembly process in which releasing one set of constraints depends on the motion permitted by the other.
Since the exact material properties of end-of-life products are not known a priori, the feasible disassembly direction cannot be reliably derived from CAD geometry alone. This motivates a force-driven approach that infers feasible extraction directions from real-time contact interaction rather than relying on a fixed, precomputed motion primitive.

More generally, robotic assembly and disassembly remain challenging due to contact-rich interactions and variability in part geometry, pose, and surrounding environments, a difficulty that continues to motivate dedicated contact-rich manipulation frameworks~\cite{goli2024disassembly, tang2023industreal, luo2024serl}. Furthermore, end-of-life products often exhibit wear, deformation, and structural irregularities that deviate from their nominal CAD models, limiting the reliability of purely model-based approaches~\cite{tian2022assemble, tian2024asap}, an issue compounded by geometry preprocessing pipelines that can fail to preserve fine-grained features~\cite{wei2022approximate} such as hooks. In addition, many critical connections are occluded within assemblies, reducing the effectiveness of vision-based perception, motivating recent force and tactile-based alternatives that recover task-relevant contact information under occlusion~\cite{he2025foar, saka2026contact, hegeler2025teaching}.

\begin{figure}
    \centering
    \includegraphics[width=\linewidth]{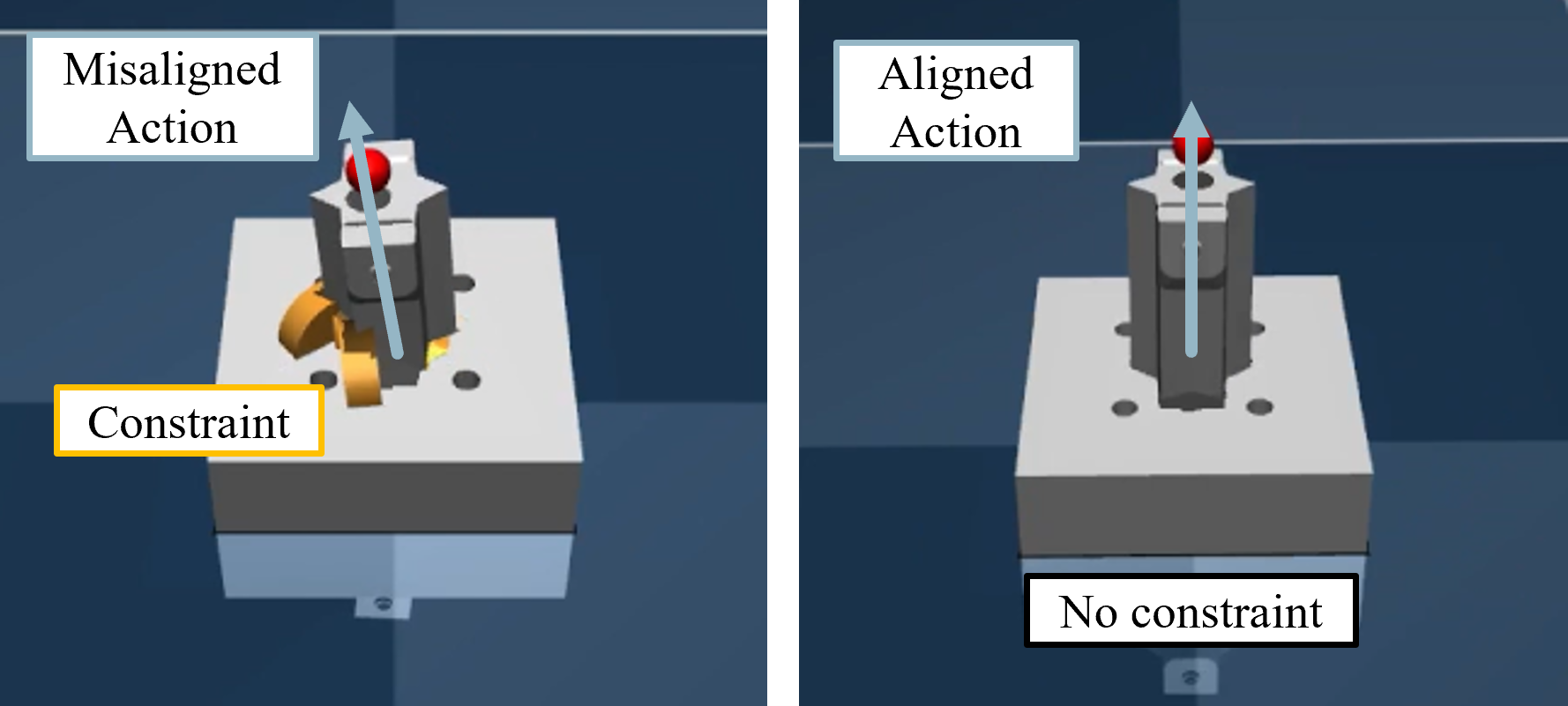}
    \caption{Training Environment: Simulated contact interaction under misaligned and aligned extraction actions. When the commanded action direction (cyan arrow) deviates from the groundtruth extraction axis (left), the peg engages the surrounding constraint geometry, producing visible contact forces (gold contact markers). On the other hand, when the action direction is more aligned to the groundtruth axis (right), contact decreases until it disappears at complete alignment. This illustrates that contact forces carry direct information about the feasible motion direction, which the proposed method exploits during learning.}
    \label{fig:trainingparts_direction}
\end{figure}

Existing methods typically rely on explicit geometric reasoning or predefined motion primitives to guide disassembly. Although effective in structured settings, these approaches struggle to capture the complex and evolving constraints present in tightly coupled assemblies, particularly when small but critical features or contact interactions are difficult to model accurately.
In contrast, this work leverages contact-rich interaction as a source of informative sensory feedback rather than a challenge to avoid. In tightly constrained assemblies, contact forces implicitly encode information about feasible motion directions and underlying structural constraints as illustrated in Fig.~\ref{fig:trainingparts_direction}. By incorporating this feedback into a reinforcement learning framework, the proposed method learns continuous disassembly policies that infer feasible extraction strategies directly from interaction, without requiring predefined joint types or explicit geometric sequences.

Industrial disassembly of such products therefore requires addressing three challenges: (1) inferring feasible disassembly directions under unknown and partially observable constraints, (2) exploiting contact interactions to achieve stable and non-destructive manipulation, and (3) learning policies that generalize across diverse products without product-specific modeling or reconfiguration.

To address (1) and (2), we learn extraction behavior from contact forces in a physics-based simulation that provides rich contact feedback while avoiding repeated exploratory contact that EoL parts cannot sustain. To address (3), the policy is trained on a simplified peg-in-hole pair in a robot agnostic, object centric environment, so that it learns a force following behavior based on interaction dynamics rather than product-specific geometry or a particular robot. However, instantaneous contact force signals are inherently noisy and locally ambiguous, particularly for multi-stage joints where the feasible axis changes as the connection transitions between constraint states. A policy conditioned only on momentary force feedback can therefore act inconsistently, producing jitter and redundant exploratory motion (quantified in Section~\ref{sec: results} via direction reversals and path efficiency). To address this, we couple the policy with a particle filter (PF) belief estimator that integrates contact observations over time into a persistent belief over the feasible axis, giving the policy a temporally consistent signal to condition its actions on, rather than merely reacting to instantaneous force readings. In this work, we refer to the feasible axis as the joint axis for consistency with previous work.

The contributions of this paper are twofold:
\begin{enumerate}
    \item We propose a constraint-aware, object-centric RL framework for force-driven hook disassembly, trained on a simplified peg-in-hole pair in a robot-agnostic simulation environment, that transfers zero-shot to a physical robot and to unseen sequential constraint structures without product-specific models or demonstrations.
    \item To address the noise and local ambiguity of instantaneous force feedback, we introduce a particle filter-based belief over the feasible extraction direction, integrated into the RL policy through observation augmentation and reward shaping. Across four real-world tasks, the belief-augmented policy achieves the highest path efficiency on every task and raises the success rate on a real EoL air-conditioning cover with occluded, multi-stage hook-and-lip constraints from $45\%$ to $75\%$.
\end{enumerate}

Experimental results demonstrate that the proposed constraint-aware learning approach enables efficient and non-destructive disassembly of complex hook-type connections while reducing reliance on detailed geometric modeling and task-specific planning. Moreover, because training relies on a generic peg-in-hole model, the framework avoids the need for high-fidelity, watertight assembly models during training in physics engines, thereby significantly reducing modeling effort and mitigating errors introduced by geometric decomposition.

The remainder of this paper is organized as follows. Section II reviews related work and identifies assumptions of existing approaches that EoL disassembly violates. Section III presents the framework. Section IV describes the experimental design, implementation, and metrics, and Section V reports results. Section VI discusses failure modes, limitations, and future work, and Section VII concludes this work.

\section{Related Work}
\subsection{Model- and Planning-Based Approaches}

The increasing demand for remanufacturing end-of-life (EoL) products has motivated extensive research into automated disassembly planning and manipulation. \cite{tian2022assemble} propose a physics-based assembly-by-disassembly planner using an SDF-based rigid body simulator to explore feasible disassembly paths, and \cite{tian2024asap} extend this framework by incorporating gravitational stability and robotic execution constraints, performing disassembly planning through tree search guided by geometric heuristics or a graph neural network while a physics-based stability checker determines the required grasp supports for stable execution. 
Other model-based approaches infer explicit structural representations from sensor data and use them to guide classical controllers. \cite{hegeler2025teaching} learn kinematic constraint models from passive RGB-D demonstrations by segmenting object trajectories into contact modes and fitting parameterized geometric constraints for online contact detection and force control through an admittance controller. \cite{SIMONIC2024102657} propose a hierarchical framework in which a robot discovers contact mode transitions through physical exploration using a Constraint-Space-Following controller, incrementally constructing a graph of feasible contact modes while a Q-learning agent learns optimal traversal strategies.

These approaches rely on accurate geometric models or externally observable task structure. EoL products, however, frequently exhibit wear, deformation, and manufacturing variance that render watertight CAD representations incomplete~\cite{tian2022assemble, tian2024asap}, and standard geometry preprocessing such as convex decomposition can fail to preserve fine-grained features such as hook connections~\cite{wei2022approximate}, which are critical to the disassembly constraint structure. Similarly, exploration-based approaches such as~\cite{SIMONIC2024102657} require the constraint manifold dimensionality and feasible probing directions to be specified by the user, which is not possible when the fastening geometry is internally occluded and its degrees of freedom are unknown in advance.

\subsection{Learning from Demonstration-Based Approaches}
Imitation learning methods learn manipulation policies directly from demonstrations, providing a flexible alternative to explicit model construction. \cite{he2025foar} propose a force-aware reactive policy that combines force/torque sensing with point cloud observations through a future contact predictor. \cite{saka2026contact} investigate tactile sensing for robotic disassembly using a visuotactile imitation learning framework, showing that structured force-based representations outperform raw tactile imagery, particularly under tight contact constraints. Hierarchical approaches decompose contact-rich manipulation into multiple levels of abstraction, separating high-level task sequencing from low-level motion execution to improve sample efficiency and generalization. \cite{Wang2021} decouple trajectory planning from force control by combining goal-conditioned imitation learning with reinforcement learning-based force adaptation, and \cite{sun25b} combine predefined low-level manipulation primitives with a high-level policy for sequencing and parameterization in long-horizon assembly.
 
These methods assume that demonstrations can be collected for, and transfer across, product variants. Beyond the practical cost of collecting sufficient teleoperated demonstrations~\cite{he2025foar, saka2026contact, hegeler2025teaching}, slight variations in fastening geometry can induce discontinuous changes in feasible disengagement trajectories, making behavior cloning~\cite{Wang2021, sun25b} brittle regardless of demonstration quantity. Moreover, when fastening mechanisms are embedded within housings, the feasible disengagement direction cannot be inferred from external appearance alone; as the results of~\cite{saka2026contact} indicate, force responses carry task-relevant information that visual observations cannot provide under occluded contact constraints, so physical interaction is required to reveal the constraint structure.

\subsection{Reinforcement Learning-Based Approaches}
Reinforcement learning (RL) enables policies to adapt through interaction rather than relying on predefined motion plans or demonstrations, and has been applied to contact-rich assembly and disassembly with tight tolerances, uncertain dynamics, and sim-to-real transfer. Several approaches integrate compliance and force awareness: \cite{goli2024disassembly} apply SAC (Soft Actor Critic) with adaptive compliance control to peg-hole disassembly, showing that force-aware RL can mitigate jamming and wedging during extraction, and \cite{zhang2024peg} employ SAC for cylindrical peg-hole disassembly using force-torque feedback, achieving more stable and lower-force extraction than DDPG (Deep Deterministic Policy Gradient). Other works improve training and system design: \cite{tang2023industreal} combine simulation-aware updates, geometry-informed rewards, and curriculum learning to improve the robustness and transferability of insertion, \cite{luo2024serl} develop a real-world RL system emphasizing safe impedance control and efficient online learning, and \cite{Ota2019} use an RRT-generated reference path to guide trajectory learning through reward shaping. Abstraction has also been explored to improve scalability and transfer: \cite{li2025flow} show that force-aware visuomotor policies can generalize from minimal simulation demonstrations through compliant control and force-informed data augmentation, \cite{saito2023constraint} organize policies by physical constraint type, such as prismatic or revolute, enabling reuse across tasks, \cite{fang2025flex} train force-based policies directly in object force space without robot-specific kinematics, and \cite{thierauf2024fixing} advocate learning in object-centric rather than robot-centric spaces. Finally, \cite{zang2025robotic} combine reward shaping with a rule-based residual control policy derived from disassembly heuristics, improving learning efficiency and transferability across clearance-fit disassembly tasks and highlighting the benefits of integrating prior task knowledge into RL-based skill acquisition.

Two assumptions of these works limit their use in EoL disassembly. First, contact-rich learning on real hardware requires that components sustain repeated exploratory interactions or that safe exploration boundaries be defined in advance, neither of which holds when parts are non-replaceable and the constraint structure is unknown until contact reveals it; this motivates simulation-based training with sim-to-real transfer, as in~\cite{goli2024disassembly, zhang2024peg, tang2023industreal} where a policy can develop force-following behavior across diverse and geometrically irregular constraint structures without risking component damage, before being deployed on real hardware. 
Second, several methods rely on prior knowledge of the extraction structure, such as a reference path~\cite{Ota2019} or a known extraction direction assumed by the residual policy of~\cite{zang2025robotic}. While effective for clearance-fit assemblies with predefined kinematic structure, such assumptions are difficult to satisfy in EoL products whose fastening geometries are hidden, damaged, or involve multiple sequential constraint stages, where the feasible direction must instead be inferred online from interaction.

\subsection{Positioning of This Work}
Taken together, these limitations motivate a framework that (i) infers constraint directions online from contact forces rather than from models, vision, or demonstrations, (ii) is trained in simulation to avoid damaging non-replaceable parts, and (iii) is expressed in object space so that it does not depend on a specific target product or robot. Although a policy conditioned on a short window of recent observations can infer the feasible direction implicitly, such inference remains noisy and prone to direction reversals. We therefore complement it with an explicit probabilistic belief that maintains uncertainty over the latent extraction direction and accumulates evidence over the whole interaction as contact reveals the underlying constraint structure.

\section{Proposed Framework}
\subsection{Overview}
\label{sec:overview}
\begin{figure*}[tb] 
    \centering \includegraphics[width=\textwidth]{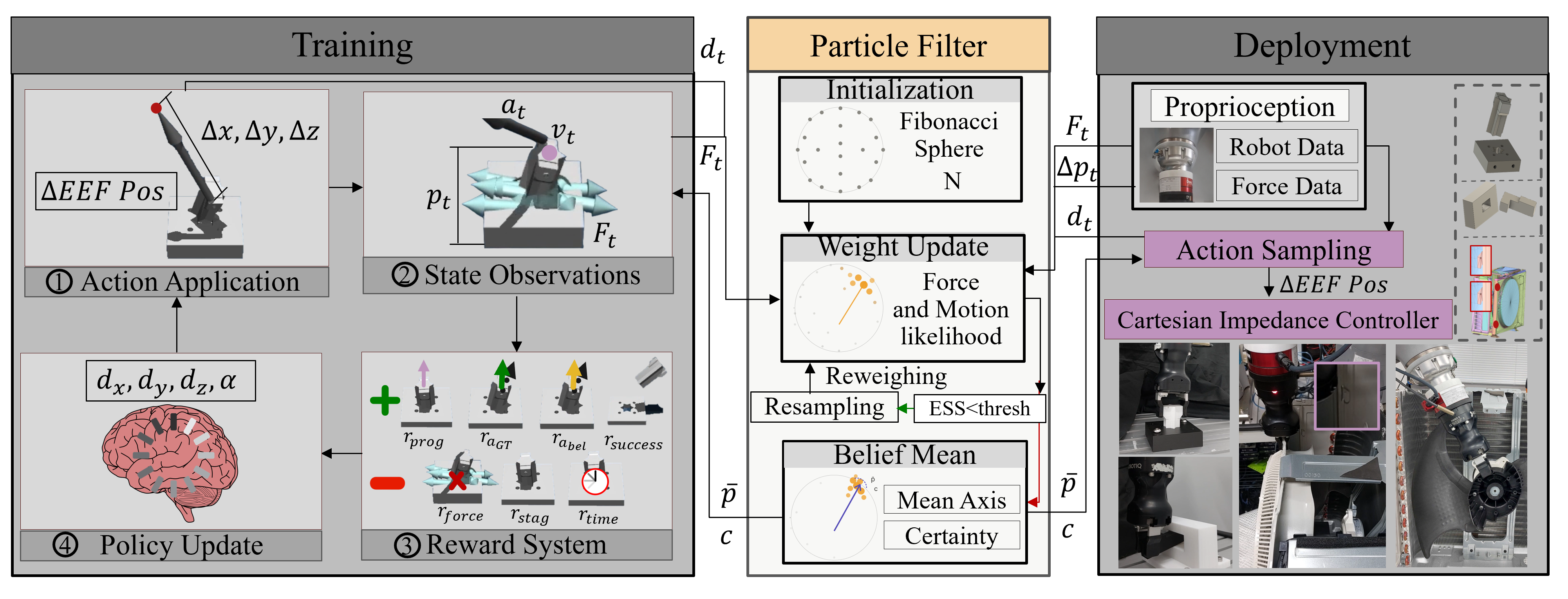} \caption{Overview of the proposed framework.
        \textbf{Left:} During training, the SAC policy receives state observations and belief estimates and is shaped by a multi-component reward.
        \textbf{Center:} The particle filter maintains a belief over the feasible extraction direction, updated from force-motion observations (Section~\ref{sec:pf}). 
        \textbf{Right:} During deployment, a single-system condition is adopted where the gripper and grasped object are assumed to move as a rigid composite body and the wrist-measured forces equal the constraint force on the object \cite{saito2023constraint}. Using proprioceptive information, force sensor data, and particle filter belief specifics, the policy drives a Cartesian impedance controller on the physical robot.} 
    \label{fig:approachwithPF} 
\end{figure*}

The proposed framework illustrated in Fig.~\ref{fig:approachwithPF} represents the disassembly process as a long-horizon peg-in-hole extraction task. During both training and deployment, the system combines two parallel components: a particle filter that maintains an online belief $(\bar{\mathbf{p}}, c)$ over the locally feasible extraction direction from force-motion information, and a reinforcement learning policy that generates extraction actions $a_t = [d_x, d_y, d_z, \alpha]^T$ and applies them in the form of displacement commands $\Delta x = \alpha \hat{\mathbf{d}}_t$ applied directly to the peg. To form a closed-loop, the policy receives a temporal stack of $K$ recent observations of the environment together with $(\bar{\mathbf{p}}, c)$, where belief informs action selection, and the resulting force-motion observations are in-turn used to update the belief.

\subsection{Problem Setting}
We formulated hook disassembly as a constrained extraction problem in which a component must be removed through interaction with an unknown geometric constraint structure. As the component progresses through the extraction process, the locally feasible motion direction may vary due to changing contact conditions and geometric interactions. Rather than assuming access to the true disassembly direction, the task was modeled as a partially observable decision-making problem in which the feasible extraction direction must be inferred from force-motion interactions. This formulation was selected in training to emulate the uncertainty encountered during real disassembly operations, where the underlying constraint structure and feasible removal directions are typically unknown a priori.

Although the training geometry is a peg-in-hole pair, the policy is trained only for extraction. Unlike insertion, which terminates in a fully constrained state at the bottom of the hole, extraction always admits a locally unconstrained direction until the part is free, which makes following the least-constrained direction a well-posed strategy for disassembly. Moreover, fasteners such as hooks and lips are released through a sequence of motions along different directions rather than by reversing a single insertion motion.

A key assumption of the proposed approach is that, despite potentially complex contact interactions, local disassembly progress can be approximated by a feasible dominant motion direction. Although hook connections may exhibit locally varying constraints and deviations from a single extraction axis, successful disassembly generally requires maintaining motion within a subset of directions that produce low resistance and positive extraction progress. Consequently, the objective of the agent is not to reproduce a predefined trajectory but rather to continuously infer and exploit feasible motion directions from observed force responses. 

\subsection{Particle-Based Belief Representation}
\label{sec:pf}
Because the policy generates actions from short-horizon stacked observations, its behavior can be locally noisy and prone to exaggerated direction reversals, as it has no explicit mechanism for summarizing evidence accumulated beyond a short temporal window. To address this, an explicit belief over the locally feasible extraction direction is maintained throughout execution, rather than relying on an implicit internal state. This belief summarizes the accumulated evidence from previous force-motion interactions and is supplied to the policy together with a scalar confidence measure, allowing action selection to depend on a temporally-integrated estimate rather than solely on instantaneous observations.

Particle filters provide a non-parametric framework for maintaining probabilistic beliefs over latent states and have been widely applied in robotics problems involving uncertainty and partial observability. Their ability to represent arbitrary non-Gaussian distributions over a manifold-constrained state makes them particularly suitable when the underlying state cannot be directly measured and must instead be inferred from indirect observations. In contact-rich manipulation, particle-filter-based approaches have been used to estimate geometric and contact properties from force measurements~\cite{kutsuzawa2026simultaneous}. Similarly, particle filters have been integrated with reinforcement learning to maintain explicit belief states in partially observable environments, improving decision-making under uncertainty~\cite{ma2020discriminative}.

For this work, a particle filter is employed to maintain an online belief over the locally feasible disassembly direction. The same filter structure and belief output are used during training and deployment, while the rules by which observations update the belief differ between the two. Because the RL policy only observes the filter through its output, the update rules can be adapted to the physical system without retraining the policy. The shared representation is described first, followed by the observation models, the training-time update, and the deployment-time update. All parameter values are listed in Table~\ref{tab:pf_params}.

\textit{Belief representation and output.}
The belief over the extraction direction is represented by a set of $N$ weighted unit vectors,
\begin{equation}
\mathcal{P}=\{(\mathbf{p}_i, w_i)\}_{i=1}^{N}, \qquad \|\mathbf{p}_i\|=1, \qquad \textstyle\sum_i w_i = 1,
\end{equation}
where each particle $\mathbf{p}_i$ is a candidate hypothesis for the feasible direction. At the start of every attempt, particles are initialized on a Fibonacci lattice, which distributes points quasi-uniformly over the sphere and avoids any initial clustering near the poles that can result from naive latitude-longitude sampling, with uniform weights $w_i = 1/N$ reflecting the absence of prior information about the feasible direction.

The belief supplied to the policy is the weighted mean direction together with a scalar confidence given by the mean resultant length of the particle distribution:
\begin{equation}
\bar{\mathbf{p}} = \frac{\sum_i w_i \mathbf{p}_i}{\left\|\sum_i w_i \mathbf{p}_i\right\|}, \qquad c = \Big\|\textstyle\sum_i w_i \mathbf{p}_i\Big\| \in [0,1],
\label{eq:belief_output}
\end{equation}
where $c \to 0$ when the particles are dispersed over the sphere and $c \to 1$ when they are concentrated around a single direction. The pair $(\bar{\mathbf{p}}, c)$ is concatenated with the environment observations and supplied to the policy network in both training and deployment.

\textit{Force observation.} At each update, the filter uses the commanded motion direction
\begin{equation}
\hat{\mathbf{d}}_t = \frac{[d_x,d_y,d_z]^\top}{\|[d_x,d_y,d_z]\|},
\end{equation}
obtained from the directional component of the policy action (Section~\ref{sec:overview}), and the contact force it produced. The force vector is smoothed with an exponential filter, 
\begin{equation}
\mathbf{F}_t = \alpha_F \mathbf{F}_t^{\text{raw}} + (1 - \alpha_F)\mathbf{F}_{t-1}, \qquad \alpha_F=0.4,
\end{equation}
to suppress transient force spikes while remaining responsive to genuine changes in resistance, and its magnitude is normalized to a scalar force response $f_t \in [0,1]$, where $f_t = 0$ indicates free motion and $f_t = 1$ indicates a fully blocked push; the normalization differs between training and deployment and is given below. Each particle predicts the response that a push along $\hat{\mathbf{d}}_t$ would produce if $\mathbf{p}_i$ were the feasible direction. The alignment model
\begin{equation}
\hat{f}^{\,a}_i = \frac{1-\mathbf{p}_i^\top \hat{\mathbf{d}}_t}{2}
\label{eq:f_align}
\end{equation}
assigns low resistance to motions aligned with the feasible direction, high resistance to motions opposing it, and intermediate resistance to sideways motion. A stricter physical model,
\begin{equation}
\hat{f}^{\,p}_i = \sqrt{1-\max\!\big(0,\, \mathbf{p}_i^\top\hat{\mathbf{d}}_t\big)^2},
\label{eq:f_phys}
\end{equation}
treats only pushes along the feasible direction as unresisted and both sideways pushes and pushes into the constraint as blocked. The force likelihood of a particle is evaluated from the discrepancy between the predicted and observed responses,
\begin{equation}
g(\hat{f}) = (1-\varepsilon)\exp\!\left(-\frac{(\hat{f}-f_t)^2}{2\sigma^2}\right) + \varepsilon,
\label{eq:force_lik}
\end{equation}
where $\sigma$ controls the sensitivity of the update and the floor $\varepsilon$ prevents a single anomalous reading from eliminating a particle.

\textit{Motion observation.} When the part moves, the direction of its displacement is direct evidence of the feasible direction. Given the measured displacement direction $\hat{\mathbf{v}}$, the motion likelihood
\begin{equation}
L^{\text{mot}}_i = (1-\varepsilon)\exp\!\big(\kappa\,(\mathbf{p}_i^\top \hat{\mathbf{v}} - 1)\big) + \varepsilon
\label{eq:motion_lik}
\end{equation}
favors particles aligned with the observed motion, with $\kappa$ controlling how sharply they are preferred.

\textit{Training-time update.}
During training, the filter is updated once per control step from force observations only, following every policy action. The force response is normalized by a fixed saturation constant, $f_t = \min(\|\mathbf{F}_t\|, F_{\max})/F_{\max}$, the likelihood uses the alignment model without a floor, $L_i = g(\hat{f}^{\,a}_i)$ with $\varepsilon = 0$, and the weights are updated multiplicatively,
\begin{equation}
w_i \leftarrow \frac{w_i L_i}{\sum_j w_j L_j}.
\end{equation}
To reduce weight degeneracy, the weights are then blended with a uniform distribution, 
\begin{equation}
    w_i \leftarrow (1-\eta)\,w_i + \eta/N, \qquad \eta = 10^{-4}.
\end{equation}
To prevent particle impoverishment, systematic resampling is performed when the effective sample size $N_{\mathrm{eff}} = 1/\sum_i w_i^2$ falls below $N/2$, followed by regularized rejuvenation, in which each resampled particle is perturbed with isotropic Gaussian noise of fixed standard deviation $h_0 = (4/(3N))^{1/6}$ and renormalized to the unit sphere. The particle count $N = 64$ was selected empirically to balance angular resolution against the computational cost of updating the filter at every control step.

\textit{Deployment-time update.}
On the physical system, two properties of real force sensing break the assumptions of the training-time update. Under the compliant impedance controller, a blocked push produces only a few newtons of resistance rather than the large constraint forces of simulated contact, so a fixed normalization by $F_{\max}$ makes nearly every push appear free and lets the belief simply follow the policy's own actions. 
In addition, once the part moves, its measured displacement provides direct evidence of the feasible direction that the force signal alone does not capture. 
As in training, force updates are applied once per policy action, using the mean of the last three force readings of that action after discarding the initial transient, so that correlated readings of the same push are not counted as independent evidence. 
The deployed filter differs from the training-time filter in four respects.

\textit{(i) Evidence sources and gating.} Motion and force observations are both used, each when it is informative. A motion update with likelihood~(\ref{eq:motion_lik}) is performed whenever the end-effector has moved by at least $\Delta_{\min}$ since the previous motion update, using positions averaged over the last three samples to suppress vibration. A force update is performed only when the part moved less than $1$~mm during the action, since the force is then informative about the constraint, whereas during motion the drag force would contradict the motion evidence; it is further skipped when $\|\mathbf{F}_t\| < F_\text{c}$, since a push that meets no resistance carries no information about the constraint.

\textit{(ii) Force normalization and model.} The force response is normalized by the resistance actually encountered rather than by $F_{\max}$,
\begin{equation}
f_t = \min\!\left(\frac{\max(0,\, \|\mathbf{F}_t\| - F_\text{db})}{F_\text{ref}},\, 1\right),
\end{equation}
where $F_\text{db}$ is a sensor-noise dead-band and $F_\text{ref} = \min\!\big(\max(F_\text{ref,min}, F_\text{peak}), F_{\max}\big)$ follows the largest resistance $F_\text{peak}$ seen so far in the episode, which decays by a factor of $0.98$ per update. Since it is not known in advance whether real contact behaves more like~(\ref{eq:f_align}) or~(\ref{eq:f_phys}), the force likelihood averages the two,
\begin{equation}
L^{\text{for}}_i = \tfrac{1}{2}\big(g(\hat{f}^{\,a}_i) + g(\hat{f}^{\,p}_i)\big),
\end{equation}
with a floor $\varepsilon > 0$ in~(\ref{eq:force_lik}).

\textit{(iii) Tempered update with forgetting.} Both evidence sources update the weights in log form,
\begin{equation}
\log w_i \leftarrow \lambda \log w_i + \tau \log L_i, \qquad w_i \leftarrow \frac{\exp(\log w_i)}{\sum_j \exp(\log w_j)},
\label{eq:tempered_update}
\end{equation}
where $L_i$ is $L^{\text{for}}_i$ or $L^{\text{mot}}_i$. The exponent $\tau < 1$ accounts for the residual correlation between successive readings, and the forgetting factor $\lambda < 1$ lets old evidence decay, so that the belief can follow the feasible direction while it changes between constraint stages.

\textit{(iv) Resampling and recovery.} Resampling is triggered when $N_{\mathrm{eff}} < 0.3N$ rather than $N/2$, so that accumulated evidence is not discarded after few updates. The rejuvenation noise is scaled to the current spread of the particle cloud, 
\begin{equation}
h = \mathrm{clip}\big(\sigma_\text{cloud} N^{-1/6},\, 1.5^\circ,\, 12^\circ\big)
\end{equation}
where $\sigma_\text{cloud}$ is the weighted root-mean-square angular deviation of the particles from $\bar{\mathbf{p}}$ divided by $\sqrt{2}$, and the noise is applied tangentially to the sphere, so that a concentrated belief is not dispersed by a fixed perturbation. To recover from a wrong belief, the filter tracks fast and slow running averages $\bar{L}_\text{fast}$ and $\bar{L}_\text{slow}$, of its predictive likelihood $\sum_i w_i L_i$. Following augmented Monte Carlo localization~\cite{thrun2002probabilistic}, a fraction $\min\!\big(\max(0,\, 1-\bar{L}_\text{fast}/\bar{L}_\text{slow}),\, 0.25\big)$ of the particles is uniformly redrawn from the sphere at each resampling. Particles are therefore re-injected only when the belief persistently fails to predict new observations. Finally, the particle count is increased to $N = 256$ for finer angular resolution, which remains affordable at one force update per action.

\begin{table}[t]
\centering
\caption{Particle filter parameters used during training and deployment.}
\label{tab:pf_params}
\footnotesize
\setlength{\tabcolsep}{3pt}
\resizebox{\columnwidth}{!}{%
\begin{tabular}{lcc}
\toprule
\textbf{Parameter} & \textbf{Training} & \textbf{Deployment} \\
\midrule
Particles $N$                              & $64$                  & $256$ \\
Force update rate                          & per action            & per action \\
Evidence                                   & force                 & force + motion \\
Force model                                & $\hat{f}^{\,a}$       & mean of $\hat{f}^{\,a}$, $\hat{f}^{\,p}$ \\
Force normalization                        & fixed, $F_{\max}=200$~N & adaptive, $F_\text{ref,min}=15$~N \\
Dead-band $F_\text{db}$                    & --                    & $0.5$~N \\
Contact threshold $F_\text{c}$             & --                    & $1.5$~N \\
Likelihood width $\sigma$                  & $0.3$                 & $0.2$ \\
Likelihood floor $\varepsilon$             & $0$                   & $0.05$ \\
Motion step $\Delta_{\min}$                & --                    & $1.5$~mm \\
Motion sharpness $\kappa$                  & --                    & $4$ \\
Temper $\tau$ / forgetting $\lambda$       & $1$ / $1$             & $0.7$ / $0.97$ \\
Uniform weight blend $\eta$                & $10^{-4}$             & -- \\
Resampling threshold                       & $N_\text{eff}<N/2$    & $N_\text{eff}<0.3N$ \\
Rejuvenation noise                         & $h_0\approx0.52$ (fixed) & $1.5^\circ$--$12^\circ$ (adaptive) \\
Re-injection fraction                      & --                    & $\le 0.25$ (adaptive) \\
Recovery averages $\alpha_\text{fast}$ / $\alpha_\text{slow}$ & -- & $0.2$ / $0.02$ \\
\bottomrule
\end{tabular}}
\end{table}

\subsection{Reinforcement Learning Formulation}
The reinforcement learning policy receives observations reflecting the consequent state of the environment concatenated with the current belief estimate. A temporal stack of recent observations is maintained to provide short-term interaction history and improve observability of evolving contact conditions, allowing the policy to reason over force-motion histories rather than instantaneous measurements alone. 

The disassembly task is formulated as a partially observable Markov decision process (POMDP), in which the agent does not have direct access to the locally feasible extraction direction but must infer it from accumulated force-motion observations. The policy is trained using SAC~\cite{haarnoja2018soft}, an off-policy maximum-entropy algorithm well-suited to continuous contact-rich domains. Domain randomization is applied at the start of each episode: the joint axis is sampled uniformly from the three canonical axes, the geometry is rotated accordingly, and small positional perturbations are applied to the initial configuration, encouraging the policy to learn extraction behaviors that generalize across orientations rather than overfitting to a fixed coordinate frame.

At each decision step, the policy outputs an action $\mathbf{a}_t = [d_x, d_y, d_z, \alpha]^\top \in [-1, 1]^4$, where $\mathbf{d} = [d_x, d_y, d_z]^\top$ specifies a motion direction and $\alpha$ represents the displacement magnitude. The direction vector is normalized prior to execution, $\hat{\mathbf{d}} = \mathbf{d} / \|\mathbf{d}\|$, and $\alpha$ is linearly mapped to a step size in $[0, \delta_{\max}]$, yielding the displacement command:
\begin{equation}
    \Delta \mathbf{x} = \alpha \hat{\mathbf{d}}.
    \label{eq:action}
\end{equation}
This decouples direction from magnitude, allowing the policy to independently adapt both the extraction direction and step aggressiveness in response to evolving contact conditions. The resulting displacement is applied directly to a fixed point on the top surface of the peg within the object-centric simulation environment, hereafter referred to as the end-effector. Applying actions in object space decouples the learned behavior from a particular robot's kinematics~\cite{saito2023constraint, fang2025flex, thierauf2024fixing}.

Since the task is formulated without visual observations, the state of the policy is constructed from proprioceptive and force-based measurements that capture the progress of extraction and contact interactions. The state component comprises the position of the end-effector relative to the hole site $\mathbf{p}_\text{rel} \in \mathbb{R}^3$, the velocity of the end-effector $\dot{\mathbf{p}} \in \mathbb{R}^3$, the exponentially smoothed contact force vector $\mathbf{f} \in \mathbb{R}^3$ and its scalar magnitude, the displacement magnitude of the extraction $\|\mathbf{p}_\text{rel}\| \in \mathbb{R}$ as a proxy of cumulative extraction progress, and the most recently applied action $\mathbf{a}_{t} \in \mathbb{R}^4$, totaling 15 dimensions. The belief estimate $(\bar{\mathbf{p}}, c) \in \mathbb{R}^4$ is appended to form a 19-dimensional single-step observation. To improve observability of evolving contact dynamics, a temporal stack of the $K = 12$ most recent observations is maintained, forming the full policy input $\mathbf{o}_t \in \mathbb{R}^{228}$.

The reward function jointly encourages extraction progress, penalizes excessive contact forces, and promotes consistency between actions and the current belief. The primary term is an efficiency-weighted progress signal:
\begin{equation}
    r_\text{prog} =
    \begin{cases}
        k_\text{prog} \dfrac{\delta_{\parallel}}{1 + \beta \|\mathbf{F}_t\|} & \delta_{\parallel} > 0 \\[6pt]
        -k_\text{prog} \, |\delta_{\parallel}| & \delta_{\parallel} \le 0
    \end{cases}
    \label{eq:progress_reward}
\end{equation}

where $\delta_{\parallel} = \Delta\mathbf{x}_t^\top \hat{\mathbf{z}}$ is the displacement projected onto the joint axis $\hat{\mathbf{z}}$ and $\beta$ is a force-weighting coefficient, and $k_\text{prog}$ is a scaling coefficient. When motion progresses along the joint axis, this component rewards forward progress in proportion to how efficiently it is achieved under contact load. When motion is against the joint axis, $r_\text{prog}$ is negative and scales with the magnitude of the backward displacement, discouraging regression away from the extraction direction by penalizing backward motion symmetrically. This design choice was intended to handle the middle steps of the task where both the extraction and the insertion directions are feasible with minimal forces.

A directional alignment bonus 
\begin{equation} 
    r_{a_{GT}} = k_\text{GT} \cos\theta 
    \label{eq:gt_bonus}
\end{equation}
where $\theta$ is the angle between the executed motion and the joint axis, reinforces axis-consistent behavior during training. When the particle filter confidence exceeds a threshold $c > c_\text{th}$, a belief-alignment bonus
\begin{equation}
    r_{a_{bel}} =
    \begin{cases}
        k_\text{bel} \, c \left(\mathbf{d}_t^\top \bar{\mathbf{p}}\right) & c > c_\text{th} \\[4pt]
        0 & \text{otherwise}
    \end{cases}
    \label{eq:belief_reward}
\end{equation}

additionally rewards actions consistent with the current belief estimate, coupling inference and action selection. 
Contact forces are further discouraged through a quadratic penalty with an additional hard penalty above a safe threshold $F_\text{thresh}$:

\begin{equation}
    r_\text{force} = \lambda_1 \left(\frac{\|\mathbf{F}_t\|}{F_\text{max}}\right)^{\!2} + \lambda_2 \max\!\left(0,\, \|\mathbf{F}_t\| - F_\text{thresh}\right)
    \label{eq:force_reward}
\end{equation}

Additionally, $r_\text{stag}$ is a stagnation penalty that discourages episodes in which the agent fails to make extraction progress over a sliding window of $M$ steps.
A constant time penalty $r_\text{time}$ encourages efficient completion. Upon successful disassembly, a terminal bonus $r_\text{success}$ provides a strong positive signal. The total reward is:
\begin{equation}
    r_t = r_\text{prog}  + r_{a_{GT}} + r_{a_{bel}}- r_\text{force} - r_\text{stag} -r_\text{time} + r_\text{success} 
    \label{eq:total_reward}
\end{equation}

Reward coefficients were determined empirically through iterative adjustment: an initial value was assigned to reflect the intended relative priority of each term (e.g., extraction progress prioritized over force minimization, with contact force terms scaled to remain subordinate to progress under typical operating forces), and values were then refined by qualitatively inspecting the resulting policy behavior such as the extraction ability, contact force magnitudes, and stagnation or reversal frequency, until the policy exhibited the intended trade-off between progress and safety. This trial-and-error tuning process is consistent with standard practice in reward-shaped reinforcement learning, where the reward function encodes a designer's prioritization of task objectives into a form the agent can optimize directly. Final coefficient values used in all reported experiments are listed in Table~\ref{tab:reward_summary}.

\begin{table*}[tb]
\centering
\caption{Summary of reward components used during reinforcement learning.}
\label{tab:reward_summary}
\footnotesize
\begin{tabularx}{\textwidth}{l c X c l}
\toprule
\textbf{Term} & \textbf{Sign} & \textbf{Purpose} & \textbf{GT} & \textbf{Value} \\
\midrule
$r_\text{prog}$     & $\pm$ & Encourage efficient extraction progress; penalize regression. & \checkmark & $k_\text{prog}=200,\ \beta=0.05$ \\
$r_{a_{GT}}$   & $+$   & Encourage motion along the true joint axis. & \checkmark & $k_\text{GT}=2.0$ \\
$r_{a_{bel}}$  & $+$   & Reward actions aligned with the particle-filter belief. & & $k_\text{bel}=2.0,\ c_\text{th}=0.6$ \\
$r_\text{force}$    & $-$   & Penalize excessive contact forces. & & $\lambda_1=0.5,\ \lambda_2=0.01,\ F_\text{thresh}=60\text{N}$ \\
$r_\text{stag}$     & $-$   & Penalize stagnation over multiple steps. & & $k_\text{stag}=0.01,\ M=20$ steps \\
$r_\text{time}$     & $-$   & Encourage shorter episodes. & & $0.02$ \\
$r_\text{success}$  & $+$   & Reward successful task completion. & & $20$ \\
\bottomrule
\end{tabularx}
\end{table*}

Successful disassembly is defined by two simultaneously satisfied conditions: the net constraint force projected onto the joint axis must remain negligible for $K_\text{term}$ consecutive time steps, and the cumulative displacement along the joint axis must exceed a predefined threshold $d_\text{th}$. These conditions jointly ensure that termination reflects genuine mechanical disengagement rather than transient force reductions.

Note that the ground-truth joint axis $\hat{\mathbf{z}}$ is available only during training and is used in the computation of $r_\text{prog}$ and $r_{a_{GT}}$. During deployment, the policy relies solely on force observations and the particle-filter belief estimate $(\bar{\mathbf{p}}, c)$ to guide action selection.

The proposed framework combines force-guided exploration, online direction inference, and shaped reinforcement learning rewards to identify feasible extraction motions that adapt to changing constraint conditions without requiring explicit geometric knowledge or predefined disassembly trajectories.

\section{Experiments}
\label{sec: experiments}

\subsection{Overview}
\label{exp: overview}

The experiments in this section are designed to answer four questions, the first three of increasing difficulty: \begin{enumerate}
    \item Can a policy trained entirely in simulation transfer zero-shot to physical hardware?
    \item Can a policy trained on single-axis extraction generalize to sequential multi-axis constraint structures?
    \item Can a policy trained on a single simplified connection generalize to real end-of-life industrial components with occluded, non-uniform constraints?
    \item Does the particle filter belief contribute beyond the policy's implicit direction inference, and how does the learned approach compare with a deterministic force-following strategy?
\end{enumerate}

Section~\ref{sec: results} reports five experiments addressing these questions: a single-axis sim-to-real transfer experiment on a geometrically complex star-shaped peg-in-hole pair (Section~\ref{results: results_singleaxis}), a generalization experiment on a previously unseen sequential two-axis extraction pair (Section~\ref{results: results_seqaxis}), two industrial deployments on a real end-of-life outdoor air-conditioning unit, namely extraction of the inner propeller (Section~\ref{results: results_propeller}) and removal of its front cover (Section~\ref{results: results_aircon}), and an ablation study and baseline comparison isolating the contribution of the particle filter belief representation and comparing against a deterministic force-following controller across all four real-world tasks (Section~\ref{results: results_ablation}).

Section~\ref{exp: implem} describes the training procedure of the reinforcement learning policy, policy implementation, and experimental hardware. Section~\ref{exp: eval_met} defines the quantitative metrics used to evaluate all experiments.

\subsection{System Implementation}
\label{exp: implem}
The policy was trained entirely in simulation using the MuJoCo physics engine in a robot-agnostic, object-centric environment at 100 Hz. As shown in Fig~\ref{fig:trainingparts_direction}, the training assembly is an openly available peg-in-hole model shared by the MuJoCo community. The peg and hole share a star-shaped cross-section with tight tolerances ($0.55$~mm), a design choice motivated by two complementary properties. First, the multi-lobed geometry produces distributed contact points along the extraction axis rather than localized edge contacts, generating rich and directionally informative force feedback throughout the extraction trajectory. Second, the constraint structure remains geometrically simple: successful disassembly requires only prismatic motion along a single axis, providing a clean ground-truth signal for reward shaping without introducing confounding multi-stage transitions during training. Together, these properties encourage the policy to learn force-following behavior from interaction rather than memorizing a specific geometric extraction path, which we hypothesize is the primary mechanism enabling generalization to structurally dissimilar hook connections at deployment.

Robot kinematics, gripper dynamics, and visual observations were not included during training; the policy receives only proprioceptive and force-based inputs. The policy was trained using the reference CleanRL implementation of the Soft Actor-Critic (SAC) algorithm~\cite{huang2022cleanrl}, with environment-specific modifications. All hyperparameters and network weights are fixed after training and carried over to deployment without any fine-tuning or domain adaptation.

The experiments tested the disassembly of: 1) a star-shaped peg and hole pair, 2) a sequential 2-dimensional peg and hole pair, 3) the propeller mounted on the fan shaft of an outdoor air-conditioning unit, and 4) the front cover of that unit. All experiments were conducted using a KUKA LBR IIWA 14 robot arm equipped with a wrist-mounted force-torque sensor. The policy outputs a four-dimensional action $\mathbf{a}_t = [d_x, d_y, d_z, \alpha]^\top$ which is mapped to a Cartesian displacement command executed through a Cartesian impedance controller, whose compliance keeps contact forces bounded when the commanded direction is misaligned with the constraint. The policy commands translational displacements only, which are converted to joint targets by a damped least-squares inverse kinematics solver on the peg-tip position. Therefore, the end-effector orientation is not commanded and may drift under contact loads. Unlike rotation-symmetric pegs, multi-edged cross-sections such as the star require precise rotational alignment with the hole, and even small rotational deviations cause the lobes to wedge against the hole walls~\cite{geiger2026diffusion}.
As a result, for the star-shaped assembly, an orientation-hold term was added to the inverse kinematics for all three controllers (RL + PF, RL-only, and FF) where at each step, $50\%$ of the rotation error with respect to the initial end-effector orientation is corrected, limited to $1^\circ$ per step and weighted at $0.2$~m/rad relative to the position error. This term acts only on orientation and does not alter the translational commands of any controller.

The Cartesian impedance controller runs on the robot controller with translational and rotational stiffness of $4000$~N/m and $300$~N$\cdot$m/rad and a damping ratio of $0.7$ in all experiments. Its tool center point (TCP), about which the impedance is defined and which therefore acts as the compliance center, is placed at the tip of the grasped part, since locating the compliance center at the end of the extracted part minimizes jamming and extraction force~\cite{goli2024characterizing}.

Because the framework uses no visual perception, a small set of task-specific quantities was set manually before each experiment and kept fixed across all trials and controllers: the TCP offset from the flange, the tool load (mass and center of mass of the gripper and grasped part) used by the controller for gravity compensation, the reference point from which the relative position $\mathbf{p}_\text{rel}$ is computed, and the displacement threshold used to end an episode.

The single-system condition~\cite{saito2023constraint} is adopted throughout deployment: the robot arm's end-effector and the grasped object move as a single rigid composite body. Under this condition, internal forces between the gripper and object cancel, and the wrist-measured force directly equals the constraint force acting on the object, allowing the same force-based observations to be used without a gripper contact model.
This condition is satisfied in practice by using a Robotiq Hand-E parallel gripper mounted on the robot flange, consistent with industrial disassembly settings, and tightly grasping the removable object between the gripper fingers to ensure a stable grasp prior to initiating extraction.
For this setup, the grasp location is selected manually prior to each trial, prioritizing grasp feasibility and stability. For the peg-in-hole pairs, the peg is grasped at its exposed upper section. However, for the outdoor air-conditioning unit, the geometric detailing of the front cover prevents grasping at its center. Since the cover is retained by two downward hooks on one side and a continuous lip mechanism along the opposite edge, the gripper is positioned closer to the hook side, where structural rigidity is sufficient to maintain the single-system condition during extraction. However, this placement transmits little load to the lip-side fastener farthest from the grasp, which resulted in partial disengagement in several trials, as discussed in Section~\ref{sec: limitations}. For the propeller, the gripper was angled to avoid collision with the part of the outer frame that sits directly above it.

The particle filter and policy run online alongside the robot controller using only wrist force measurements and proprioceptive data, with no access to visual input or geometric models of the target assembly.

\subsection{Evaluation Metrics}
\label{exp: eval_met}
Five quantitative metrics are reported across experiments. The first metric is the \textbf{success rate}, defined as the proportion of trials in which full extraction is achieved without breaking or permanently deforming the target connection. For the air-conditioning unit experiments, a trial is considered successful only if the front cover is fully disengaged from the fixed frame and the hook connection remains structurally intact after removal; trials in which the cover was partially disassembled while a fastener remained engaged are counted as failures. For the propeller, a trial is considered successful once the propeller has been displaced along the shaft beyond a predetermined threshold. A trial is considered a failure when the predefined maximum duration of $50$ s was exceeded in the task. The second metric is the \textbf{mean task completion time} over successful trials. To characterize the constraint-following behavior under contact, the \textbf{average peak constraint force} is additionally reported. Two further metrics characterize how extraction was achieved: the \textbf{path efficiency}, the ratio of the net displacement to the actual executed path length where a higher percentage indicates more direct extraction and less jitter, and the \textbf{direction reversal count}, which quantifies back-and-forth translational motion during extraction through a count of sign changes between consecutive samples of each Cartesian velocity component, summed across the three translational axes. A higher count indicates more frequent changes in translational motion and is used as an indicator of exploratory or jittery extraction behavior. These five metrics are applied identically across all four real-world tasks and are summarized in Table~\ref{tab:realworld_results} and Table~\ref{tab:policy_comparison}.

\section{Results}
\label{sec: results}
\subsection{Single-Axis Extraction: Sim-to-Real Transfer}
\label{results: results_singleaxis}

The first real-world experiment evaluates whether the policy trained entirely in simulation transfers directly to physical hardware without any fine-tuning or additional training. The target assembly is a geometrically complex star peg-in-hole whose constraint structure was included in the training distribution, isolating the sim-to-real transfer question. 

\begin{figure}[t]
    \centering
    \includegraphics[width=0.8\linewidth, height=0.4\textwidth]{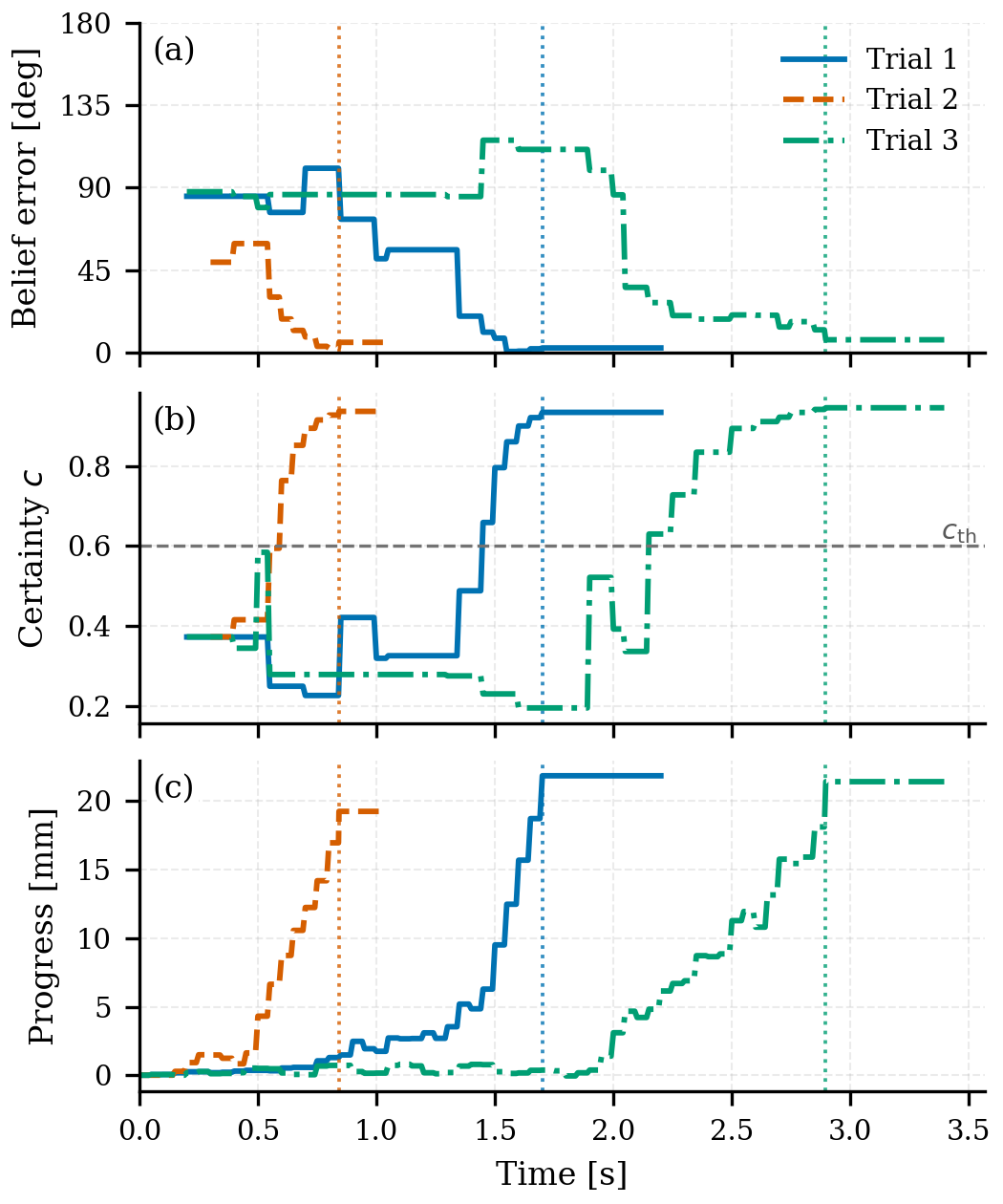}
    \caption{Particle filter belief evolution on the star-shaped peg-in-hole task for three representative trials (fast, typical, and slow). (a) Angle between the belief $\bar{\mathbf{p}}$ and the true extraction axis. (b) Belief confidence $c$, with the reward threshold $c_\text{th}$ dashed. (c) Displacement along the true extraction axis. Dotted vertical lines mark task completion. In each trial, extraction accelerates as the belief error falls and the confidence rises, while early phases with an inaccurate belief produce little progress.}
    \label{fig:pf_convergence}
\end{figure}

The robot arm is placed at its start position where the grasped peg is initiated inside the fixed hole. The policy is deployed zero-shot by converting the sampled 4-dimensional action to a 3-dimensional difference in position that is fed into the Cartesian impedance controller. Over 20 trials, the system achieved a success rate of  $100\%$ with a mean task completion time of $1.98~\mathrm{s}$ over successful trials, an average peak force of $25.65~\mathrm{N}$, a path efficiency of $47\%$, and an average of $49.2$ direction reversals. Because the true extraction axis of this assembly is known, it is also used to evaluate the particle filter belief directly; Fig.~\ref{fig:pf_convergence} shows the angular error between the belief and the true axis, the belief confidence, and the extraction progress over time for three representative trials.

\subsection{Generalization to Multi-Stage Constraints: Sequential Two-Axis Disassembly}
\label{results: results_seqaxis}

\begin{figure}[t]
    \centering
    \includegraphics[width=\linewidth]{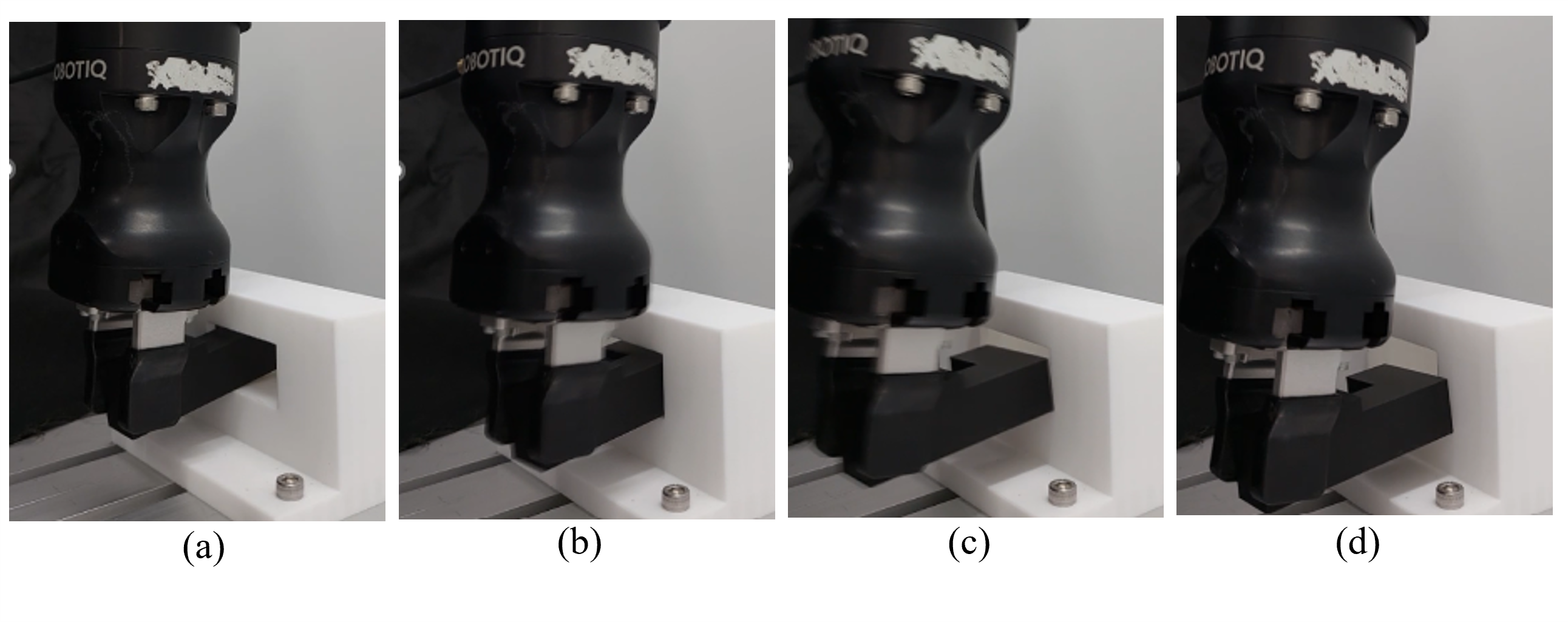}
    \caption{Sequential two-axis disassembly task. From left to right: (a) initial insertion configuration, (b) transition point where the extraction direction changes, (c) continued extraction along the second orthogonal axis, and (d) complete removal of the peg from the fixture.}
    \label{fig:sequentialpeg_disassemblysteps}
\end{figure}

The second experiment evaluates whether the constraint-following skill generalizes to assemblies with more complex constraint structures than those seen during training. The target is a sequential two-axis assembly in which the component must first be displaced along an initial prismatic axis before transitioning to a second orthogonal axis to complete disassembly. No information about the axes or the constraint structure is provided to the policy at deployment. Fig.~\ref{fig:sequentialpeg_disassemblysteps} illustrates the execution sequence of the real-world task.

Over 20 trials, the system achieved a success rate of $100\%$ with a mean task completion time of $2.07~\mathrm{s}$, a mean peak contact force of $33.14~\mathrm{N}$, a path efficiency of $82\%$, the highest of the four tasks, and an average of $17.4$ direction reversals.  

\subsection{Industrial Deployment: Inner Propeller of an EoL Air-Conditioning Unit}
\label{results: results_propeller}
The third experiment extracts the propeller mounted on the fan shaft beneath the front cover of the air-conditioning unit, as shown in Fig.~\ref{fig:approachwithPF} (right); the cover was removed beforehand. Although the propeller is removed along a single axis, the specific shaft orientation as well as friction from dirt and rust on the worn shaft makes the required extraction force unknown a priori, and no information about the shaft axis is provided to the policy.
The propeller was grasped at a feasible point and extracted by the policy. Note that the tightening bolt was removed by hand before this experiment, since bolt removal is out of the scope of this work.

A total of 20 trials were conducted and achieved a success rate of $90\%$ with a mean task completion time of $8.41~\mathrm{s}$, an average peak force of $36.84~\mathrm{N}$, a path efficiency of $29\%$, and an average of $138.89$ direction reversals. Both failed trials ended in a timeout after the propeller had been partially displaced along the shaft. In these trials, the propeller's free rotation about the shaft offered low resistance to tangential motion, so lateral displacements produced force responses similar to those of the feasible extraction direction. Because the belief and the action space represent translational directions only, this unmodeled rotational degree of freedom made the force signal locally ambiguous: the belief shifted toward tangential directions, and the policy continued moving without further progress along the shaft until the time limit was reached. This failure mode is discussed further in Section~\ref{results: failure_modes}.

Table~\ref{tab:realworld_results} summarizes the results of the proposed method across all four real-world tasks, which are interpreted in Section~\ref{results: failure_modes}.

\begin{table*}[t]
    \footnotesize
    \centering
    \caption{Real-world results of the proposed method (RL + PF) across all four tasks.}
    \label{tab:realworld_results}
    \begin{tabular}{lcccccc}
    \toprule
    \textbf{Task} & \textbf{Trials} & \textbf{Success [\%]} & \textbf{Time [s]} & \textbf{Peak Force [N]} & \textbf{Path Eff. [\%]} & \textbf{Direction Reversals} \\
    \midrule
    Star Peg-in-Hole     & 20 & 100 & 1.98  & 25.65 & 47 & 49.2   \\
    Sequential Two-Axis  & 20 & 100 & 2.07  & 33.14 & 82 & 17.4   \\
    Propeller            & 20 & 90  & 8.41  & 36.84 & 29 & 138.89 \\
    Front Cover          & 20 & 75  & 19.52 & 26.13 & 31 & 261.87 \\
    \bottomrule
    \end{tabular}
\end{table*}

\subsection{Industrial Deployment: Front Cover of an EoL Air-Conditioning Unit}
\label{results: results_aircon}
The fourth experiment evaluates deployment on a real end-of-life outdoor air-conditioning unit, which presents multiple non-trivial hook connections with unknown geometry, manufacturing variance, and surface wear. This experiment represents the most demanding test of the framework, as the constraint structure is internally occluded, non-uniform across units, and subject to deformation-induced variation that cannot be modeled in advance.

\begin{figure}[t]
    \centering
    \includegraphics[width=\linewidth]{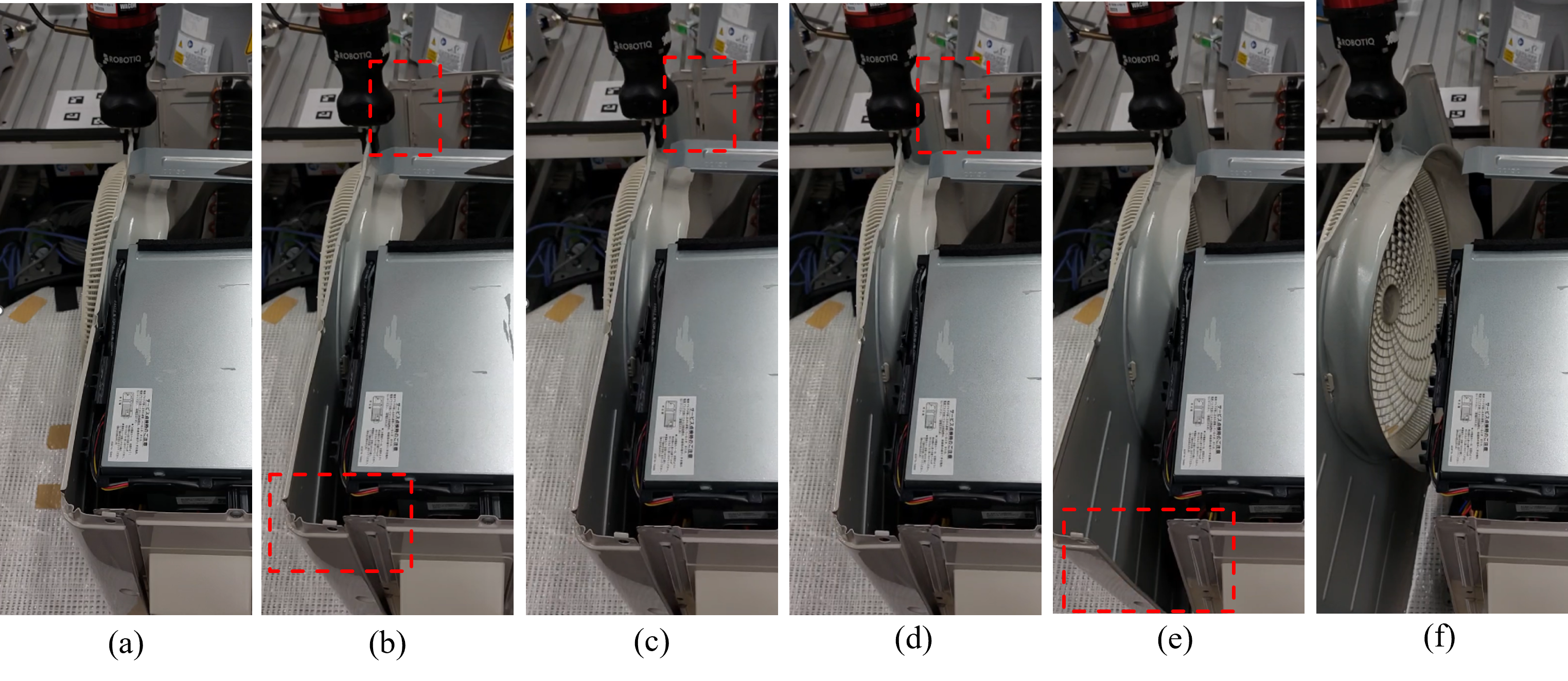}
    \caption{EoL air-conditioning unit front cover disassembly task. From left to right: (a) initial insertion configuration, (b) initial upward motion, (c) hook release, (d) distance increase, (e) full lip disengagement, and (f) full disassembly.}
    \label{fig:AirConDisassemblySequence}
\end{figure}

A total of 20 trials were conducted and achieved a success rate of $75\%$ with a mean task completion time of $19.52~\mathrm{s}$, an average peak force of $26.13~\mathrm{N}$, a path efficiency of $31\%$, and an average of $261.87$ direction reversals. Fig.~\ref{fig:AirConDisassemblySequence} illustrates the execution sequence of the real-world task. Of the five failed trials, three ended in partial disengagement, in which the cover was released from the hooks while the bottom-right fastener, farthest from the grasp point, remained engaged, and two stalled with no further motion after the initial upward release. Partial disengagement also occurred under the RL-only and FF controllers, indicating that it stems from the grasp location rather than from the controller; this is examined in Section~\ref{results: failure_modes}.

\subsection{Ablation Study and Baseline Comparison}
\label{results: results_ablation}
To isolate the contribution of the particle filter belief representation, we compare the full proposed method against a variant in which the particle filter outputs $(\bar{\mathbf{p}}, c)$ are removed from the observation and the belief-alignment reward $r_{a_{bel}}$ is disabled during training. This ablation is conducted on real hardware across all four real-world tasks.
In this ablated variant, the policy receives only proprioceptive and force observations and must infer extraction directions without an explicit belief estimate.

To assess whether a deterministic rule acting on the same force signal suffices, we additionally evaluate a learning-free force-following (FF) baseline designed for this study. FF implements a four-phase intuitive strategy of probing for the least-constrained direction, committing to it while it yields progress, and searching anew when motion stalls. In the initialization phase, the mean force over the first few control steps is recorded as a bias $\mathbf{f}_0$. In the probing phase, FF commands short displacements from a common origin along $14$ fixed directions (the six signed Cartesian axes and the eight body diagonals), retracting to the origin after each. A probe ends when the resistance along the probe direction, $-(\mathbf{f}-\mathbf{f}_0)^\top\mathbf{d}$, builds up to $F_\text{low}=3$~N, when the part has moved $6$~mm, or after $8$ steps, and is scored as $\min(F_\text{peak},F_\text{high})/F_\text{high} + (1-\Delta/6~\text{mm})$, with $F_\text{high}=30$~N and $\Delta$ the achieved displacement, such that lower scores indicate freer directions. In the motion phase, FF moves along the best-ranked direction $\mathbf{h}$ with step $\delta = \delta_\text{max}(1-0.7\,s)$, where $s=\mathrm{clip}\big((\|\mathbf{f}-\mathbf{f}_0\|-F_\text{low})/(F_\text{high}-F_\text{low}),0,1\big)$ is the normalized load. When the load exceeds $F_\text{low}$, the heading is steered along the component of the force perpendicular to $\mathbf{h}$ with gain $0.6\,s$, or by a perpendicular sidestep when the block is head-on. If the distance from the start position has not increased by more than $1$~mm for $13$ steps, FF switches to the next-ranked direction, skipping directions within $26^\circ$ of the last six tried, and re-enters the probing phase from its current position once the ranked list is exhausted. FF runs through the same deployment pipeline, Cartesian impedance controller, grasp, success criteria, force smoothing ($\alpha_F=0.4$), and $\delta_\text{max}=5$~mm step bound as the learned policies, and the same action smoothing is applied during its motion phase, so that only the decision-making component differs between controllers. Comparative results can be seen in Table~\ref{tab:policy_comparison}.

\begin{table*}[t]
\centering
\begin{threeparttable}
\caption{Comparison of the full method (RL + PF) with the RL-only ablation and the force-following (FF) baseline across all real-world tasks. Bold indicates the best value per task.}
\label{tab:policy_comparison}
\footnotesize
\begin{tabular}{llcccccc}
\toprule
\textbf{Policy} & \textbf{Task} & \textbf{Trials} & \textbf{Success [\%]} $\uparrow$ & \textbf{Peak Force [N]} $\downarrow$ & \textbf{Time [s]} $\downarrow$ & \textbf{Path Eff. [\%]} $\uparrow$ & \textbf{Direction Reversals} $\downarrow$ \\
\midrule
\multirow{4}{*}{RL + PF (ours)}
 & Star Peg-in-Hole     & 20 & \textbf{100} & \textbf{25.65} & \textbf{1.98}  & \textbf{47} & \textbf{49.2}  \\
 & Sequential Two-Axis  & 20 & \textbf{100} & \textbf{33.14} & \textbf{2.07}  & \textbf{82} & \textbf{17.4}  \\
 & Propeller            & 20 & 90           & 36.84          & 8.41           & \textbf{29} & 138.89         \\
 & Front Cover          & 20 & 75           & \textbf{26.13} & 19.52          & \textbf{31} & 261.87         \\
\midrule
\multirow{4}{*}{RL Only (PF ablation)}
 & Star Peg-in-Hole     & 20 & \textbf{100} & 37.90          & 2.35           & 16          & 71.8           \\
 & Sequential Two-Axis  & 20 & 80           & 42.92          & 11.14          & 17          & 219.5          \\
 & Propeller            & 20 & \textbf{95}  & 42.24          & \textbf{5.05}  & 19          & \textbf{92.42} \\
 & Front Cover          & 20 & 45           & 31.01          & \textbf{17.75} & 14          & \textbf{242.44}\\
\midrule
\multirow{4}{*}{Force Following (FF)}
 & Star Peg-in-Hole     & 20 & \textbf{100} & 86.90          & 10.98          & 4           & 369.4          \\
 & Sequential Two-Axis  & 20 & 55           & 43.17          & 31.87          & 16          & 378.45         \\
 & Propeller            & 20 & 80           & \textbf{32.36} & 25.50          & 11          & 331.81         \\
 & Front Cover          & 20 & \textbf{90}  & 29.64          & 30.39          & 25          & 365.83         \\
\bottomrule
\end{tabular}
\begin{tablenotes}\footnotesize
\item Time is averaged over successful trials. For the front cover task, trials in which the front cover was removed but a fastener remained engaged (partial disengagement) are counted as failures.
\end{tablenotes}
\end{threeparttable}
\end{table*}

Table~\ref{tab:policy_comparison} shows that the full method is the only controller that maintains a success rate of at least $75\%$ on every task. On the two peg-in-hole tasks, it matches or exceeds both alternatives in success rate while achieving the lowest peak force, the shortest completion time, the highest path efficiency, and the fewest direction reversals; on the sequential two-axis assembly in particular, removing the belief reduces success from $100\%$ to $80\%$, increases the completion time from $2.07$ to $11.14~\mathrm{s}$, and increases direction reversals more than twelvefold. On the air-conditioning cover, the belief raises the success rate from $45\%$ to $75\%$ relative to RL-only. On the propeller, RL-only achieves a slightly higher success rate ($95\%$ vs. $90\%$), a shorter completion time, and fewer reversals, while the full method retains a higher path efficiency. 

The FF baseline succeeds on every star trial, but with more than three times the peak force of the full method ($86.90$ vs. $25.65~\mathrm{N}$) and the lowest path efficiency of all controllers, which we attribute to its repeated probing. On the sequential assembly, its success rate drops to $55\%$: in 9 of 20 trials, FF moved back and forth along the first axis after reaching the transition point instead of identifying the second axis. At the transition, motion in both directions along the first axis meets little resistance, so a rule that selects directions from the current force response alone repeatedly re-selects the axis it has already traversed. Because FF ranks candidate directions once from probes made at the start of the episode, where the second axis is still blocked, its fallback at the transition point is the reverse of the first axis, and a new probing phase is triggered only after the ranked list is exhausted.
The full method, whose belief accumulates evidence over the entire interaction and whose policy is trained to avoid regression, completed every trial on this assembly with the fewest direction reversals ($17.4$ vs. $378.45$ for FF). Finally, FF achieves the highest success rate on the air-conditioning cover ($90\%$), at the cost of the longest completion time ($30.39~\mathrm{s}$). These patterns are examined in Section~\ref{results: failure_modes}.

\section{Discussion}
\label{sec: discussion}

\subsection{Role of the Belief and Comparison with Baseline}
\label{disc: belief}
Across the four real-world tasks (Table~\ref{tab:realworld_results}), the policy does not appear to learn the geometry of the training peg. Instead, it learns a force-following strategy that seeks locally feasible directions under contact constraints, a behavior that depends on interaction dynamics rather than geometric appearance. This is consistent with its zero-shot transfer from a single-axis star peg to the sequential peg and to the cover and propeller of the air-conditioning unit, whose geometries differ substantially from the training environment. On the star assembly, where the true axis is known, the belief converged to the extraction direction in every trial, with a median final error of $5.8^\circ$ (within $10^\circ$ in $16$ of $20$ trials and below $26^\circ$ in all) and a final confidence of at least $0.88$, and extraction accelerated as its error fell and its confidence rose (Fig.~\ref{fig:pf_convergence}), indicating that the policy acts on the belief rather than ignoring it.

The comparison in Table~\ref{tab:policy_comparison} indicates that the benefit of the belief grows with the amount of evidence that must be retained over the interaction. On the sequential assembly where the feasible direction changes partway through extraction, the success rate increases from $55\%$ for FF which selects directions from its current force response and an initial ranking of probed directions, to $80\%$ for the RL-only policy which conditions on a short window of recent observations, and to $100\%$ for the full method whose belief integrates evidence over the whole interaction; direction reversals decrease correspondingly from $378.45$ to $219.5$ and $17.4$ for the three methods respectively. On the air-conditioning cover where the constraints are occluded and multi-stage, the belief raises the success rate from $45\%$ to $75\%$ relative to RL-only. In contrast, on the single-axis propeller, the RL-only policy achieved a slightly higher success rate and a shorter completion time, indicating that when a single extraction direction exists and the force signal points towards it, implicit direction inference from recent observations is sufficient, and locking on an incorrect belief can lead to failure if not recovered within the task time limit. Across all four tasks, the full method achieved the highest path efficiency, consistent with the role of the belief in reducing redundant exploratory motion.

Force-control-based strategies and the FF baseline evaluated here require no training and act directly on the measured force, which makes them effective when a single release direction exists and the contact forces point toward it: FF succeeded on every star trial and, although removing the air-conditioning cover requires a sequence of directions, achieved the highest success rate on it ($90\%$ vs.\ $75\%$). This came at the cost of repeated exploration, longer completion times ($10.98$ vs.\ $1.98$~s on the star and $30.39$ vs.\ $19.52$~s on the cover), and higher peak forces (more than three times those of the full method) on the star. The limitation of FF is clearest when the feasible direction changes after substantial progress: on the sequential peg-in-hole pair, it repeatedly re-selected the first axis at the transition point, because its candidate directions had been ranked from probes made before the second axis became free. The proposed method instead treats force as evidence about a latent feasible direction and accumulates it into a belief, which allows it to handle unknown and sequentially changing directions without relying on CAD models or visual input.

\subsection{Failure Modes and Limits of Applicability}
\label{results: failure_modes}
Three distinct failure modes were observed. First, on the air-conditioning cover, three of the five failed trials ended in partial disengagement, in which the cover was released from the hooks while the lip-side fastener farthest from the grasp (bottom right edge) remained engaged. Partial disengagement also occurred in ten RL-only trials and one FF trial, indicating that it stems from the grasp location, which transmits little load to the distant fastener, rather than from the controller. The lower rate of partial disengagement under FF with the same grasp suggests that this fastener can be released by pushing more persistently against it.
Second, the remaining two failures on the cover stalled after the initial vertical release, with the policy issuing small actions that produced little force response. This identifies a limit of applicability: performance is bounded by the informativeness of the available force signal, so poorly gripped components or configurations without a distinct resistance signature can leave the belief unable to commit to a direction. Third, on the propeller, both failures arose from the propeller's free rotation about the shaft, which offered low resistance to tangential motion. Because the belief and the action space represent translational directions only, this unmodeled rotational degree of freedom made the force signal locally ambiguous and drew the belief toward tangential directions.

\subsection{Limitations and Future Work}
\label{sec: limitations}

The grasp location on the front cover of the air-conditioning unit was selected manually and positioned near the hook-side connection due to the panel's geometric detailing. As the grasp point determines the wrench transmitted to the fasteners during extraction, it directly influences the force profiles observed by the policy and, as the partial disengagements show, the likelihood of complete disassembly. Automated grasp selection for end-of-life panels with heterogeneous detailing remains an open problem, which we intend to address using visual cues in future work. More generally, since no visual perception is used, task-specific quantities such as the TCP offset, tool load, reference position, and episode-end threshold were set manually for each product, and obtaining them automatically would be needed for fully autonomous deployment.

The belief and the action space currently cover translational directions only. Extending both to rotational degrees of freedom would allow components such as the propeller, whose release involves or permits rotation, to be represented and could remove the ambiguity that caused its failures. In addition, explicit stagnation detection could trigger deliberate probing when the belief confidence stops increasing, rather than relying on the policy to discover such recovery behavior implicitly.

Finally, the particle filter used at deployment differs from the one used during training: its update was adapted to the force magnitudes and part motion of the physical system, while the policy was not retrained. Training the policy with a filter whose observation model matches the force scale and motion evidence in simulation may further reduce the remaining gap and is left for future work.

\section{Conclusion}
This paper presented a reinforcement learning framework for contact-rich disassembly with sequentially changing constraint directions, in which a policy that infers feasible extraction directions online from contact forces is coupled with a particle filter belief that supplies temporally integrated directional information and constrains the exploration inherent to model-free RL. Across four real-world tasks, the belief-augmented policy achieved the highest path efficiency relative to a policy without belief augmentation and to a deterministic force-following baseline.

The policy was trained in a robot-agnostic simulated peg-in-hole environment and transferred zero-shot to a physical manipulator, achieving $100\%$ success on both peg-in-hole pairs, including one with a sequential two-axis constraint, and success rates of $90\%$ and $75\%$ on the inner propeller and the front cover of a real, previously unseen, end-of-life air-conditioning unit. 
These results indicate that force-driven, model-free reinforcement learning is a viable strategy for automated disassembly of end-of-life products under unknown or variable constraint conditions, without reliance on prior geometric models or demonstration data.

The framework's current applicability is bounded by its dependence on an informative force signal, its restriction to translational directions, the manual selection of grasp locations and task-specific settings, and the adaptation of the deployed belief filter without retraining the policy. Future work includes visually-guided grasp selection, rotational belief representations, stagnation-triggered probing, and training with a filter matched to real force sensing.

\section*{Funding}
This article is based on results obtained from the project JPNP23002, commissioned by the New Energy and Industrial Technology Development Organization (NEDO).

\bibliographystyle{IEEEtran}
\bibliography{paper}

@misc{balde2024global,
  title={Global e-waste monitor 2024},
  author={Bald{\'e}, Cornelis P and Kuehr, Ruediger and Yamamoto, Tales and McDonald, Rosie and D’Angelo, Elena and Althaf, Shahana and Bel, Garam and Deubzer, Otmar and Fernandez-Cubillo, Elena and Forti, Vanessa and others},
  year={2024},
  publisher={International Telecommunication Union (ITU) and United Nations Institute for Training and Research (UNITAR)}
}

@article{tian2022assemble,
  title={{Assemble Them All}: Physics-based planning for generalizable assembly by disassembly},
  author={Tian, Yunsheng and Xu, Jie and Li, Yichen and Luo, Jieliang and Sueda, Shinjiro and Li, Hui and Willis, Karl DD and Matusik, Wojciech},
  journal={ACM Transactions on Graphics (TOG)},
  volume={41},
  number={6},
  articleno={278},
  year={2022}
}

@inproceedings{tian2024asap,
  title={{ASAP}: Automated sequence planning for complex robotic assembly with physical feasibility},
  author={Tian, Yunsheng and Willis, Karl DD and Al Omari, Bassel and Luo, Jieliang and Ma, Pingchuan and Li, Yichen and Javid, Farhad and Gu, Edward and Jacob, Joshua and Sueda, Shinjiro and others},
  booktitle={Proc. IEEE ICRA},
  pages={4380--4386},
  year={2024}
}

@inproceedings{hegeler2025teaching,
  title={Teaching contact-rich tasks from visual demonstrations by constraint extraction},
  author={Hegeler, Christian and Rozzi, Filippo and Roveda, Loris and Haninger, Kevin},
  booktitle={2025 IEEE 21st International Conference on Automation Science and Engineering (CASE)},
  pages={3500--3505},
  year={2025},
  organization={IEEE}
}

@article{SIMONIC2024102657,
title = {Hierarchical learning of robotic contact policies},
journal = {Robotics and Computer-Integrated Manufacturing},
volume = {86},
pages = {102657},
year = {2024},
issn = {0736-5845},
doi = {https://doi.org/10.1016/j.rcim.2023.102657},
url = {https://www.sciencedirect.com/science/article/pii/S0736584523001321},
author = {Mihael Simonič and Aleš Ude and Bojan Nemec},
}

@article{he2025foar,
  title={Foar: Force-aware reactive policy for contact-rich robotic manipulation},
  author={He, Zihao and Fang, Hongjie and Chen, Jingjing and Fang, Hao-Shu and Lu, Cewu},
  journal={IEEE Robotics and Automation Letters},
  volume={10},
  number={6},
  pages={5625--5632},
  year={2025},
  publisher={IEEE}
}

@article{saka2026contact,
  title={CONTACT: CONtact-aware TACTile Learning for Robotic Disassembly},
  author={Saka, Yosuke and Hu, Jyun-Chi and Desai, Adeesh and Zhang, Zhiyuan and Zhang, Bihao and Luu, Quan Khanh and Prince, Md Rakibul Islam and Zheng, Minghui and She, Yu},
  journal={arXiv preprint arXiv:2603.08560},
  year={2026}
}

@INPROCEEDINGS{Ota2019,
  author={Ota, Kei and Jha, Devesh K. and Oiki, Tomoaki and Miura, Mamoru and Nammoto, Takashi and Nikovski, Daniel and Mariyama, Toshisada},
  booktitle={2019 IEEE/RSJ International Conference on Intelligent Robots and Systems (IROS)}, 
  title={Trajectory Optimization for Unknown Constrained Systems using Reinforcement Learning}, 
  year={2019},
  volume={},
  number={},
  pages={3487-3494},
  doi={10.1109/IROS40897.2019.8968010}}

@INPROCEEDINGS{Wang2021,
  author={Wang, Yan and Beltran-Hernandez, Cristian C. and Wan, Weiwei and Harada, Kensuke},
  booktitle={2021 IEEE International Conference on Robotics and Automation (ICRA)}, 
  title={Robotic Imitation of Human Assembly Skills Using Hybrid Trajectory and Force Learning}, 
  year={2021},
  volume={},
  number={},
  pages={11278-11284},
  doi={10.1109/ICRA48506.2021.9561619}}

@InProceedings{sun25b,
  title = 	 {ARCH: Hierarchical Hybrid Learning for Long-Horizon Contact-Rich Robotic Assembly},
  author =       {Sun, Jiankai and Curtis, Aidan and You, Yang and Xu, Yan and Koehle, Michael and Chen, Qianzhong and Huang, Suning and Guibas, Leonidas and Chitta, Sachin and Schwager, Mac and Li, Hui},
  booktitle = 	 {Proceedings of The 9th Conference on Robot Learning},
  pages = 	 {2628--2642},
  year = 	 {2025},
  editor = 	 {Lim, Joseph and Song, Shuran and Park, Hae-Won},
  volume = 	 {305},
  series = 	 {Proceedings of Machine Learning Research},
  month = 	 {27--30 Sep},
  publisher =    {PMLR},
  url = 	 {https://proceedings.mlr.press/v305/sun25b.html},
}

@inproceedings{goli2024disassembly,
  title={Disassembly of Rectangle Peg-Hole Structures Through Compliance Strategy and Reinforcement Learning},
  author={Goli, Farzaneh and Wang, Yongjing and Saadat, Mozafar},
  booktitle={2024 IEEE 20th International Conference on Automation Science and Engineering (CASE)},
  pages={3926--3931},
  year={2024},
  organization={IEEE}
}

@inproceedings{zhang2024peg,
  title={Peg-hole robotic disassembly compliant strategy based on soft actor-critic algorithm},
  author={Zhang, Xiaolong and Liu, Jiayi and Qi, Lei and Xu, Wenjun and Peng, Xianxu and Wu, Tianshu},
  booktitle={2024 IEEE 20th International Conference on Automation Science and Engineering (CASE)},
  pages={528--533},
  year={2024},
  organization={IEEE}
}

@inproceedings{tang2023industreal,
  author    = {Tang, Bingjie and Lin, Michael A. and Akinola, Iretiayo and Handa, Ankur and Sukhatme, Gaurav S. and Ramos, Fabio and Fox, Dieter and Narang, Yashraj},
  title     = {{IndustReal}: Transferring Contact-Rich Assembly Tasks from Simulation to Reality},
  booktitle = {Proceedings of Robotics: Science and Systems (RSS)},
  address   = {Daegu, Republic of Korea},
  year      = {2023},
  doi       = {10.15607/RSS.2023.XIX.039}
}

@inproceedings{luo2024serl,
  title={Serl: A software suite for sample-efficient robotic reinforcement learning},
  author={Luo, Jianlan and Hu, Zheyuan and Xu, Charles and Tan, You Liang and Berg, Jacob and Sharma, Archit and Schaal, Stefan and Finn, Chelsea and Gupta, Abhishek and Levine, Sergey},
  booktitle={2024 IEEE International Conference on Robotics and Automation (ICRA)},
  pages={16961--16969},
  year={2024},
  organization={IEEE}
}

@article{li2025flow,
  title={Flow with the Force Field: Learning 3D Compliant Flow Matching Policies from Force and Demonstration-Guided Simulation Data},
  author={Li, Tianyu and Li, Yihan and Zhang, Zizhe and Figueroa, Nadia},
  journal={arXiv preprint arXiv:2510.02738},
  year={2025}
}

@article{saito2023constraint,
  title={Constraint-Aware Policy for Compliant Manipulation},
  author={Saito, Daichi and Sasabuchi, Kazuhiro and Wake, Naoki and Kanehira, Atsushi and Takamatsu, Jun and Koike, Hideki and Ikeuchi, Katsushi},
  journal={Robotics},
  volume={13},
  number={1},
  pages={8},
  year={2023},
  publisher={MDPI}
}

@inproceedings{fang2025flex,
  title={FLEX: A Framework for Learning Robot-Agnostic Force-based Skills Involving Sustained Contact Object Manipulation},
  author={Fang, Shijie and Gao, Wenchang and Goel, Shivam and Thierauf, Christopher and Scheutz, Matthias and Sinapov, Jivko},
  booktitle={2025 IEEE International Conference on Robotics and Automation (ICRA)},
  pages={4782--4788},
  year={2025},
  organization={IEEE}
}

@inproceedings{thierauf2024fixing,
  title={Fixing symbolic plans with reinforcement learning in object-based action spaces},
  author={Thierauf, Christopher and Scheutz, Matthias},
  booktitle={2024 IEEE/RSJ International Conference on Intelligent Robots and Systems (IROS)},
  pages={12363--12369},
  year={2024},
  organization={IEEE}
}

@article{wei2022approximate,
  title={Approximate convex decomposition for 3d meshes with collision-aware concavity and tree search},
  author={Wei, Xinyue and Liu, Minghua and Ling, Zhan and Su, Hao},
  journal={ACM Transactions on Graphics (TOG)},
  volume={41},
  number={4},
  articleno={42},
  year={2022},
  publisher={ACM New York, NY, USA}
}

@article{kutsuzawa2026simultaneous,
  title={Simultaneous estimation of contact position and tool shape with high-dimensional parameters using force measurements and particle filtering},
  author={Kutsuzawa, Kyo and Hayashibe, Mitsuhiro},
  journal={The International Journal of Robotics Research},
  volume={45},
  number={7},
  pages={1046--1062},
  year={2026},
  publisher={SAGE Publications Sage UK: London, England}
}

@article{ma2020discriminative,
  title={Discriminative particle filter reinforcement learning for complex partial observations},
  author={Ma, Xiao and Karkus, Peter and Hsu, David and Lee, Wee Sun and Ye, Nan},
  journal={arXiv preprint arXiv:2002.09884},
  year={2020}
}

@article{haarnoja2018soft,
  title={Soft actor-critic algorithms and applications},
  author={Haarnoja, Tuomas and Zhou, Aurick and Hartikainen, Kristian and Tucker, George and Ha, Sehoon and Tan, Jie and Kumar, Vikash and Zhu, Henry and Gupta, Abhishek and Abbeel, Pieter and others},
  journal={arXiv preprint arXiv:1812.05905},
  year={2018}
}

@article{zang2025robotic,
  title={Robotic disassembly skill acquisition based on reinforcement learning with external knowledge injection},
  author={Zang, Yue and Xu, Xiazhen and Qu, Mo and Dixon, Roger and Ye, Jiaqi and Hajiyavand, Amir M and Goli, Farzaneh and Zhang, Yongquan and Wang, Yongjing},
  journal={IEEE Transactions on Industrial Informatics},
  year={2025},
  volume={21},
  number={6},
  pages={4576-4585},
  publisher={IEEE}
}

@article{huang2022cleanrl,
  title={Cleanrl: High-quality single-file implementations of deep reinforcement learning algorithms},
  author={Huang, Shengyi and Dossa, Rousslan Fernand Julien and Ye, Chang and Braga, Jeff and Chakraborty, Dipam and Mehta, Kinal and Ara{\'u}jo, Jo{\~a}o G. M},
  journal={Journal of Machine Learning Research},
  volume={23},
  number={274},
  pages={1--18},
  year={2022}
}

@article{lee2024review,
  title={A review of prospects and opportunities in disassembly with human--robot collaboration},
  author={Lee, Meng-Lun and Liang, Xiao and Hu, Boyi and Onel, Gulcan and Behdad, Sara and Zheng, Minghui},
  journal={Journal of Manufacturing Science and Engineering},
  volume={146},
  number={2},
  pages={020902},
  year={2024},
  publisher={American Society of Mechanical Engineers}
}

@article{geiger2026diffusion,
  title={Diffusion-based impedance learning for contact-rich manipulation tasks},
  author={Geiger, Noah and Asfour, Tamim and Hogan, Neville and Lachner, Johannes},
  journal={IEEE Transactions on Robotics},
  year={2026},
  publisher={IEEE}
}

@article{goli2024characterizing,
  title={Characterizing the mechanics of rectangular peg--hole disassembly and the effect of the active compliance centre on the extraction force},
  author={Goli, Farzaneh and Aflakian, Ali and Qu, Mo and Zang, Yue and Saadat, Mozafar and Pham, Duc Truong and Wang, Yongjing},
  journal={Royal Society Open Science},
  volume={11},
  number={11},
  pages={240956},
  year={2024}
}

@book{thrun2002probabilistic,
  author    = {Thrun, Sebastian and Burgard, Wolfram and Fox, Dieter},
  title     = {Probabilistic Robotics},
  publisher = {MIT Press},
  address   = {Cambridge, MA, USA},
  year      = {2005}
}

\end{document}